\documentclass{article}

\usepackage[nonatbib,final]{neurips_2021}

\usepackage[utf8]{inputenc}      
\usepackage[T1]{fontenc}         
\usepackage{hyperref}            
\usepackage{url}                 
\usepackage{booktabs}            
\usepackage{amsfonts}            
\usepackage{nicefrac}            
\usepackage{microtype}           
\usepackage{xcolor}              
\usepackage{graphicx}            
\usepackage{amsmath}             
\usepackage{algorithm}           
\usepackage{algpseudocode}       
\usepackage{caption}             

\renewcommand{\vec}[1]{\mathbf{#1}}
\newcommand{\orcid}[1]{\raisebox{1pt}{\href{https://orcid.org/#1}{\includegraphics[height=10pt]{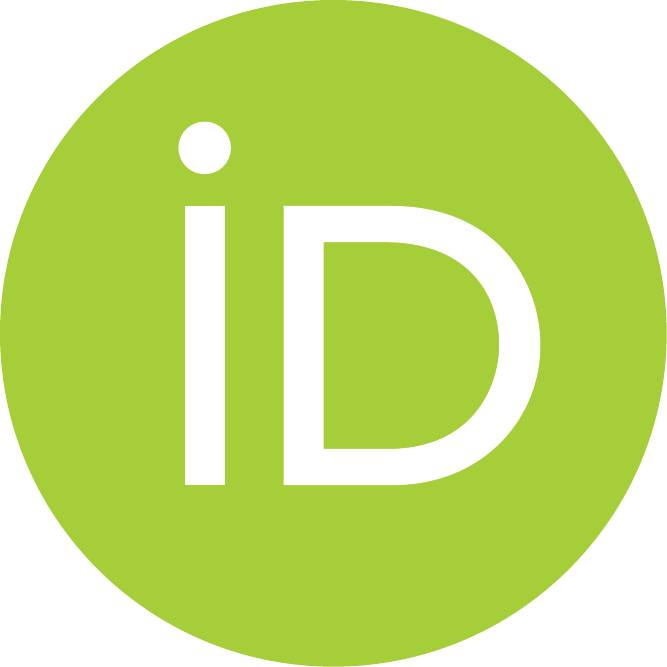}}}}
\usepackage[%
style=apa,%
backend=biber,%
citestyle=authoryear,%
natbib=true,%
uniquename=false,%
useprefix=true%
]{biblatex}
\title{Automatic Embedding of Stories Into Collections of Independent Media}

\author{%
    Dylan R.~Ashley \thanks{Correspondence to: \texttt{dylan.ashley@idsia.ch}}\enskip\textsuperscript{\rm 1,2,3} \orcid{0000-0001-6148-8802}\\
    \And
    Vincent Herrmann \textsuperscript{\rm 1,2,3}\\
    \And
    Zachary Friggstad \textsuperscript{\rm 4}\\
    \And
    Kory W. Mathewson\\
    \And
    J{\"{u}}rgen Schmidhuber \textsuperscript{\rm 1,2,3,5,6}\\
}

\begin{document}

\maketitle

\footnotetext[1]{The Swiss AI Lab IDSIA, Lugano, Switzerland}
\footnotetext[2]{Universit{\`{a}} della Svizzera italiana (USI), Lugano, Switzerland}
\footnotetext[3]{Scuola universitaria professionale della Svizzera italiana (SUPSI), Lugano, Switzerland}
\footnotetext[4]{University of Alberta, Edmonton, Canada}
\footnotetext[5]{NNAISENSE, Lugano, Switzerland}
\footnotetext[6]{King Abdullah University of Science and Technology (KAUST), Thuwal, Saudi Arabia}

\begin{abstract}

We look at how machine learning techniques that derive properties of items in a collection of independent media can be used to automatically embed stories into such collections. To do so, we use models that extract the tempo of songs to make a music playlist follow a narrative arc. Our work specifies an open-source tool that uses pre-trained neural network models to extract the global tempo of a set of raw audio files and applies these measures to create a narrative-following playlist. This tool is available at \texttt{\url{https://github.com/dylanashley/playlist-story-builder/releases/tag/v1.0.0}}

\end{abstract}

\section{Introduction}
\label{sec:introduction}

Whether it be songs, pictures, videos, or any other form of artwork, when presenting any collection of independent media, the inevitable problem arises of how the media should be ordered. In some cases, this problem is of little importance, but in many others, the overarching narrative induced by the ordering is critical to ensuring that the presentation is impactful.

If a collection of independent media is small enough---and the presentation venue important enough---then a human will often be employed to order them to produce this overarching narrative. This trend is most apparent in art galleries and museums, where a significant amount of resources have gone into the presentation of the media on display. However, the problem still exists in the same capacity when the collection is large or the venue is less important. In such situations, the ordering is often done arbitrarily for cost reasons. This work investigates how machine learning can automatically embed a story that follows a narrative arc into these orderings. While we only look at songs here, the premises we work from and many of the techniques we employ extend to any set of independent media.

\section{Automatic Embedding of Stories}
\label{sec:automatic_embedding_of_stories}

We provide evidence that albums often follow a narrative arc in Appendix~\ref{app:learning_a_predictive_scalar_representation_of_songs}. However, there are many possible narrative arcs that a collection of independent media could follow. Here, we borrow from the structure of stories as studied in \citet{freytag1894technik} and \citet{campbell2008hero} to generate a template narrative curve from the following principles: \textbf{(1)} the narrative mood should begin from a neutral point and improve slightly at first as if exposition was occurring, \textbf{(2)} the narrative mood should subsequently collapse to some negative extremum as if a crisis was occurring, \textbf{(3)} the narrative mood should then recover to a positive extremum as if a climax was occurring, \textbf{(4)} the narrative mood should finally conclude at a higher point than where it began but at a lower point than the climax, and \textbf{(5)} the exposition and conclusion should each occupy about 20\% of the total narrative.

Under the above principles, we derive the narrative template curve shown in Figure~\ref{fig:playlist_tempo}. The exact equations of this curve are provided in Appendix~\ref{app:narrative_arc_template_equation}. We use songs' global tempo---as predicted by Spotify---as a proxy for their narrative mood and use this to fit the narrative template curve through the application of the algorithm we sketch out in Figure~\ref{fig:template_curve_fitting_short}. The accompanying tool performs the above using the TempoCNN architecture of \citet{schreiber2019musical}---as provided by the Essentia library of \citet{bogdanov2013essentia}---as a means of deriving an estimate of the global tempo.

\begin{figure}[t]
    \centering
    \begin{minipage}{.575\linewidth}
        \centering
        \includegraphics[width=0.87\linewidth]{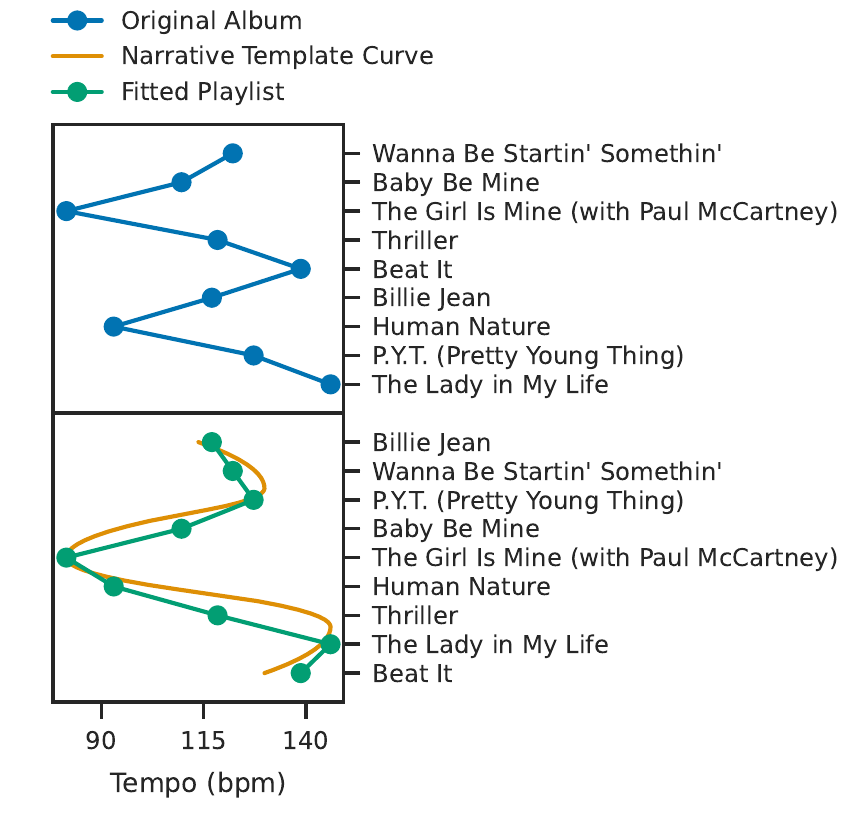}
        \captionof{figure}{The results of applying the process from Section~\ref{sec:automatic_embedding_of_stories} to the album \textit{Thriller} by Michael Jackson: the best-selling original album of all time \citep{riaa2021gold}.}
        \label{fig:playlist_tempo}
    \end{minipage}%
    \begin{minipage}{.05\linewidth}
        \hfill
    \end{minipage}
    \begin{minipage}{.35\linewidth}
        \centering
        \begin{algorithmic}[1]
            \State $f \gets$ template curve function
            \State $\vec{y} \gets$ normalized values to fit
            \State draw a bipartite graph $G$ from values to positions where edge $i \rightarrow j$ exists iff $\|\vec{y}[i] - f(\frac{j}{|\vec{y}| - 1})]\| \leq d$
            \State binary search to find $d_{min}$: the smallest possible $d$ where a perfect matching in $G$ exists
            \State draw a bipartite graph $G$ from values to positions where edge $i \rightarrow j$ with weight $w = \|\vec{y}[i] - f(\frac{j}{|\vec{y}| - 1})]\|$ exists iff $w \leq d_{min}$
            \State find a minimum cost perfect matching in $G$
        \end{algorithmic}
        \captionof{figure}{A sketch of the algorithm used to fit a playlist to the narrative arc template. The full algorithm is given in Appendix~\ref{app:template_curve_fitting_algorithm}.}
        \label{fig:template_curve_fitting_short}
    \end{minipage}
\end{figure}

\section{Related Work}
\label{sec:related_work}

The study and explicit use of narrative arcs in stories dates back over a century (e.g., \citet{freytag1894technik,vonnegut1981palm}). The use of machine learning to derive these narrative arcs has previously been explored with \citet{reagan2016emotional}---who used machine learning to explicitly derive the emotional arc of stories from a large corpus of English texts. More recently, \citet{mathewson2020shaping} used an information-theoretic approach to design a narrative arc and applied this to dialogue generation.

In contrast to the above, work in music playlist ordering remains sparse. However, a considerable body of work has recently emerged in music playlist continuation (e.g., \citet{maillet2009steerable,bonnin2013comparison,vall2019order}), and the study of form in western music has an exceptionally long and rich history with the first spatial representation that resembles a narrative arc being made by \citet{reicha1826traite}. Since then, various kinds of two-dimensional depictions have become a vital tool for the analysis and conception of musical structure or form (see \citet{bonds2010spatial}).

\section{Conclusion and Future Work}
\label{sec:conclusion_and_future_work}

We demonstrate how a story following an overarching narrative arc can be automatically embedded in a collection of independent media using machine learning-derived metrics. Future work will look at \textbf{(1)} more sophisticated machine learning techniques to describe the narrative tone of songs---such as techniques that approximate valence (we explore this briefly in Appendix~\ref{app:valence_as_a_proxy_for_narrative_mood}), \textbf{(2)} different narrative structures, \textbf{(3)} what metrics help predict the narrative structure of human-curated playlists, \textbf{(4)} alternative types of media, and \textbf{(5)} techniques to show that the induced narrative arc produces a meaningful effect for the consumer.

\section*{Acknowledgements}

This work was supported by the ERC Advanced Grant (no: 742870).

\nocite{hjelm2018learning}
\nocite{hopcroft1973n}
\nocite{jonker1987shortest}
\nocite{oord2018representation}
\nocite{oramas2016exploring}

\printbibliography

\appendix

\clearpage

\section{Learning a Predictive Scalar Representation of Songs}
\label{app:learning_a_predictive_scalar_representation_of_songs}

We used mood in our work as a feature to describe the underlying narrative of a story. However, there is no particular reason to limit ourselves to describing narratives only in terms of mood. In this section, we attempt to learn a latent representation of the underlying narratives in existing albums.  In the process of doing this, we (1) provide early evidence that existing albums have some underlying narrative structure and (2) provide a preliminary exploration of how the underlying narrative in albums can be best captured.

Here, we present a method to learn a function $e_\theta$ that maps the features $x$ of a song to a scalar $\zeta$ that maximizes the predictability of its position in an album. Our contrastive learning approach is based on deep infomax \citep{hjelm2018learning} and contrastive predictive coding \citep{oord2018representation}. It works by learning a scalar representation that maximizes the mutual information between an item and its global context. In simple terms, this means that we learn a scalar representation of a song that says the most about the global structure of the album.

To accomplish the above, we use a dataset of 11000 albums taken from the MARD dataset \citep{oramas2016exploring}. We assume that most albums are held together by some narrative structure that our learned scalar feature can capture. Each song in our dataset is represented by 13 features obtained from Spotify. In the following, $x_t$ refers to the $t$-th song in an album with a total of $T$ songs. Two LSTM networks are used to generate a global context feature vector $c_t$: network $h_\phi^f$ encodes the preceding context, and $h_\psi^b$ the subsequent context.

\begin{align*}
    \zeta_t &= e_\theta(x_t) \\
    c_t &= h_\phi^f \left(\zeta_1, \dots, \zeta_{t-1}\right) + h_\psi^b\left(\zeta_{t+1},  \dots, \zeta_{T} \right)
\end{align*}

Note that $c_t$ has no access to $\zeta_t$. A learnt function $s_\xi$ predicts $\zeta_t$ from the context $c_t$. In our setup, $s_\xi$ is a two-layer MLP with a ReLU and a sigmoid nonlinearity, and $e_\theta$ is a linear mapping followed by a sigmoid function. The system is trained to minimize the contrastive loss $\mathcal{L}$:

\begin{equation*}
    \mathcal{L}(x_t) = \left( s_\xi(c_t) - \zeta_t \right)^2 - \frac{1}{n} \sum_{i=1}^n \left( s_\xi(c_t) - e_\theta(\nu_i) \right)^2
\end{equation*}

Note that $\mathcal{L}(x_t)$ is low when the prediction $s_\xi(c_t)$ is accurate, but gets higher whenever the prediction is also close to any of the $n$ noise samples $e_\theta(\nu_i)$. It is only possible to minimize $\mathcal{L}$ if the learned scalar features $\zeta$ contain information significant to the ordering of the songs.

\begin{figure}[ht]
    \centering
    \includegraphics[width=0.4\linewidth]{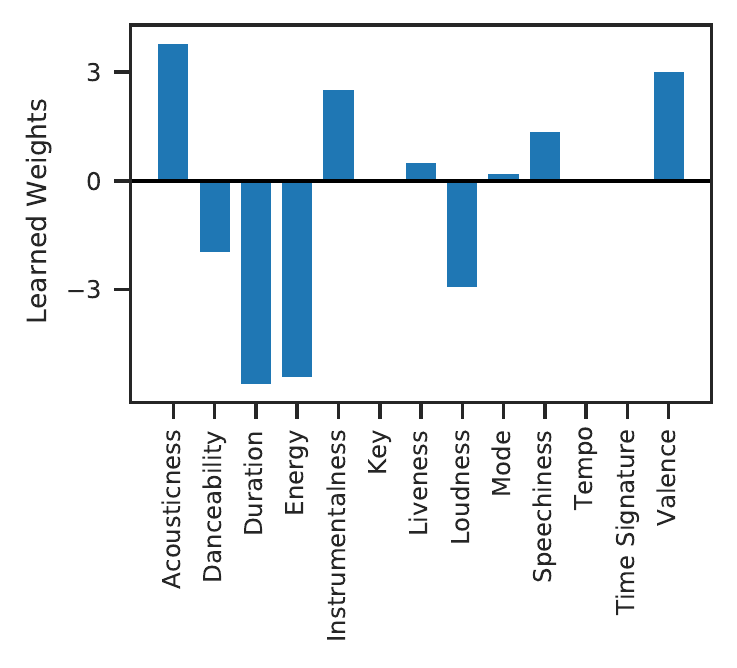}
    \captionsetup{width=0.7\linewidth}
    \captionof{figure}{Learned weights for the available features that define the scalar $\zeta$. The scalar $\zeta$ attempts to maximize the predictability of a song given the global context.}
    \label{fig:weights}
\end{figure}

With a learned scalar $\zeta$ (see Figure~\ref{fig:weights}), we achieve a validation loss of $-0.233$, which is significantly lower than with any single readily available scalar feature (energy:$-0.168$, valence:$-0.131$, duration:$-0.006$, tempo:$-0.002$). This low validation loss suggests that it does indeed capture more of the album's narrative structure than any individual feature and---consequentially---indicates that some form of a narrative structure must exist in at least some of the albums.

\begin{figure}[ht]
    \centering
    \includegraphics[width=0.5\linewidth]{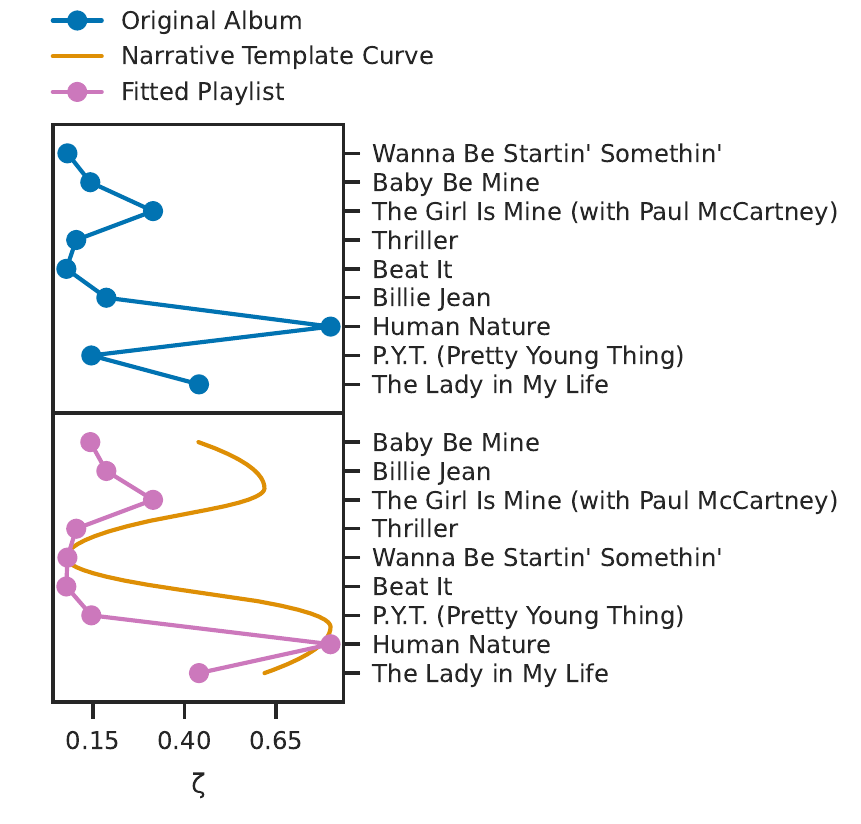}
    \captionsetup{width=0.7\linewidth}
    \captionof{figure}{The results of applying the process from Section~\ref{sec:automatic_embedding_of_stories}---using $\zeta$ instead of tempo as a proxy for the narrative mood---to the album \textit{Thriller} by Michael Jackson: the best-selling original album of all time \citep{riaa2021gold}.}
    \label{fig:playlist_zeta}
\end{figure}

Like we did with tempo in Section~\ref{sec:automatic_embedding_of_stories}, we can use $\zeta$ to embed a given story into a collection of songs, as shown in Figure~\ref{fig:playlist_zeta}. However, while we know that $\zeta$ captures some important aspects of the narrative, it is unclear to what degree this aspect correlates with the mood of a song. This means that the specific shape we introduced in Section~\ref{sec:automatic_embedding_of_stories} might not be entirely appropriate, and thus fitting $\zeta$ to it does not necessarily result in the songs following a desirable narrative arc.

\clearpage

\section{Narrative Arc Template Equation}
\label{app:narrative_arc_template_equation}

Using the principles outlined in Section~\ref{sec:automatic_embedding_of_stories}, we defined the following \((t, y)\) extremum points for a narrative arc template with \(t \in [0, 1]\) and \(y \in [0, 1]\): \((0.0, 0.5)\), \((0.2, 0.75)\), \((0.3, 0.5)\), \((0.5, 0.0)\), \((0.65, 0.5)\), \((0.8, 1.0)\), and \((1.0, 0.75)\). We then fit second order polynomials to the curve to create a piece-wise polynomial function for \(t \in [0, 1]\):
\begin{equation*}
    f(t) =
    \begin{cases}
        \frac{1}{2} + \frac{5}{2} x - \frac{25}{4} x^2 & 0.0 \leq x \leq 0.2 \\
        -\frac{1}{4} + 10 x - 25 x^2 & 0.2 < x \leq 0.3 \\
        \frac{25}{8} - \frac{25}{2} x + \frac{25}{2} x^2 & 0.3 < x \leq 0.5 \\
        \frac{50}{9} - \frac{200}{9} x + \frac{200}{9} x^2 & 0.5 < x \leq 0.65 \\
        -\frac{119}{9} + \frac{320}{9} x - \frac{200}{9} x^2 & 0.65 < x \leq 0.8 \\
        -3 + 10 x - \frac{25}{4} x^2 & 0.8 < x \leq 1.0 \\
    \end{cases}
\end{equation*}

This function is stretched along both axes to form the narrative arc template shown in Figure~\ref{fig:playlist_tempo}.

\clearpage

\section{Template Curve Fitting Algorithm}
\label{app:template_curve_fitting_algorithm}

Deriving an ordering of the media such that their respective values fit the narrative arc template given in Appendix~\ref{app:narrative_arc_template_equation} can be done using Algorithm~\ref{alg:template_curve_fitting}. The ordering Algorithm~\ref{alg:template_curve_fitting} finds will be minimal first in the maximum deviation of a value from the template curve and minimal second in the average deviation of values from the template curve. For $n$ items, the worst-case time complexity of this algorithm---provided efficient bipartite matching algorithms such as Hopcroft-Karp \citep{hopcroft1973n} and LAPJVsp \citep{jonker1987shortest} are used---is in \(O(n^{3})\). In most applications of this work---and for all but the largest collections of independent media---extracting the values that will be fitted will consume vastly more time than the fitting itself.

\begin{algorithm}[ht]
\caption{Template Curve Fitting}\label{alg:template_curve_fitting}
\hspace*{\algorithmicindent} \textbf{Input:} normalized values to fit $\vec{y}$ and template curve function $f$ with domain and range $[0, 1]$\\
\hspace*{\algorithmicindent} \textbf{Output:} ordering $\vec{x}$ over values $\vec{y}$ such that the $i$-th value in the ordering is $\vec{x}[i]$
\begin{algorithmic}[1]
\State $\vec{z} \gets \left[f(\frac{0}{|\vec{y}| - 1}), f(\frac{1}{|\vec{y}| - 1}), ..., f(\frac{|\vec{y}| - 1}{|\vec{y}| - 1})\right]^T$
\State $\vec{d} \gets \vec{y} \vec{z}^T$
\vspace{1em}
\State $a \gets 1$
\State $b \gets |\vec{d}|$
\While{$a \neq b$}
    \State $p \gets a + \lfloor (b - a) / 2 \rfloor$
    \State $L, R \gets \{1..|\vec{y}|\}$
    \State $E \gets \{(i \in L, j \in R) \mid \|\vec{y}[i] - \vec{z}[j]\| \leq \vec{d}[p]\}$
    \If{$\exists$ perfect matching for bipartite graph $(L, R, E)$}
        \State $b \gets p$
    \Else
        \State $a \gets p + 1$
    \EndIf
\EndWhile
\vspace{1em}
\State $L, R \gets \{1..|\vec{y}|\}$
\State $E \gets \{(i \in L, j \in R, \|\vec{y}[i] - \vec{z}[j]\|) \mid \|\vec{y}[i] - \vec{z}[j]\| \leq \vec{d}[a]\}$
\State $M \gets$ minimum-cost perfect matching for weighted bipartite graph $(L, R, E)$
\For{$i \in \{1..|\vec{y}|\}$}
    \For{$j \in \{1..|\vec{y}|\}$}
        \If{$(i, j) \in M$}
            \State $\vec{x}[j] = i$
        \EndIf
    \EndFor
\EndFor
\vspace{1em}
\State \Return $\vec{x}$
\end{algorithmic}
\end{algorithm}

\clearpage

\section{Valence As a Proxy for Narrative Mood}
\label{app:valence_as_a_proxy_for_narrative_mood}

\begin{figure}[ht]
    \centering
    \includegraphics[width=0.5\linewidth]{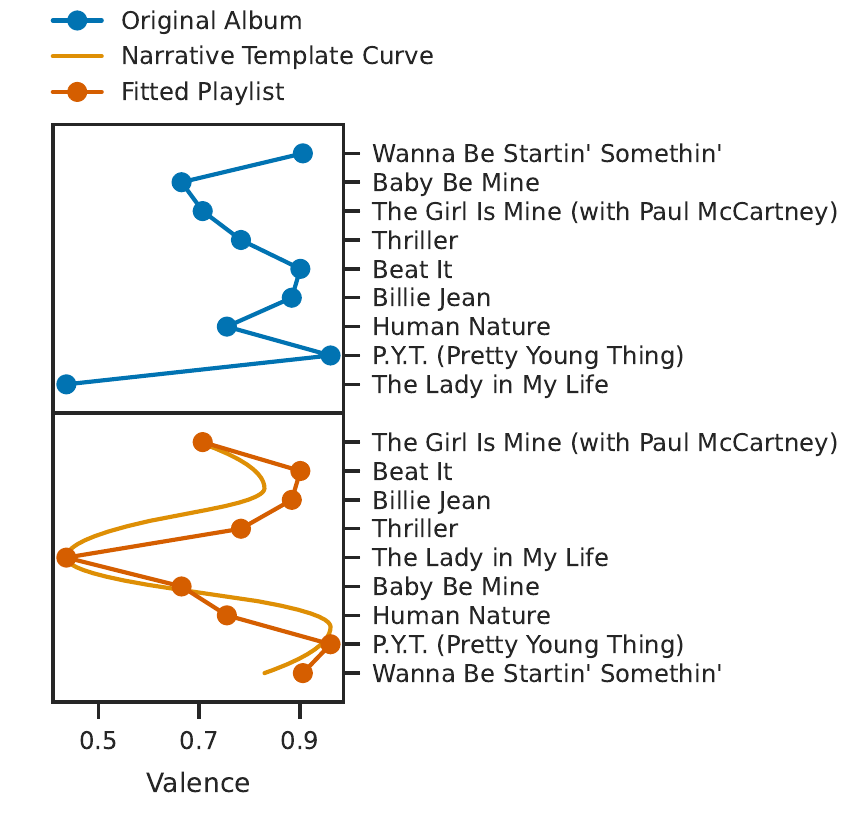}
    \captionsetup{width=0.7\linewidth}
    \captionof{figure}{The results of applying the process from Section~\ref{sec:automatic_embedding_of_stories}---using valence instead of tempo as a proxy for the narrative mood---to the album \textit{Thriller} by Michael Jackson: the best-selling original album of all time \citep{riaa2021gold}.}
    \label{fig:playlist_valence}
\end{figure}

In Section~\ref{sec:automatic_embedding_of_stories}, we used the global tempo of a song as a proxy for its narrative mood. The decision to use tempo was made because good estimates of the tempo of songs are very accessible with modern machine learning systems. However, other---potentially better---proxies exist. One particularly notable one is valence: a metric that directly attempts to capture the mood of a song. While valence is not as easily accessible as tempo, Spotify does provide estimates of it for songs in their database. Figure~\ref{fig:playlist_valence} shows the result of fitting a playlist to the narrative template curve from Appendix~\ref{app:narrative_arc_template_equation} using valence instead of tempo.

\clearpage

\section{Source Code}
\label{app:source_code}

The open-source tool described in the abstract is available at \texttt{\url{https://github.com/dylanashley/playlist-story-builder/releases/tag/v1.0.0}}

The source code used to generate the results in this experiment is available at \texttt{\url{https://gist.github.com/dylanashley/1387a99deb85bfc0bce11286810cd98b}}

\end{document}